\documentclass[runningheads]{llncs}
\usepackage{graphicx}

\usepackage{tikz}
\usepackage{comment}
\usepackage{amsmath,amssymb} 
\usepackage{color}

\usepackage[width=122mm,left=12mm,paperwidth=146mm,height=193mm,top=12mm,paperheight=217mm]{geometry}

\usepackage{multirow}
\usepackage{dsfont}
\usepackage{colortbl}
\usepackage{xcolor}
\usepackage{xspace}
\newcommand{\mtnet}{AMT\xspace}

\usepackage{cite}

\makeatletter
\DeclareRobustCommand\onedot{\futurelet\@let@token\@onedot}
\def\@onedot{\ifx\@let@token.\else.\null\fi\xspace}

\def\eg{\emph{e.g}\onedot} 
\def\ie{\emph{i.e}\onedot}

\def\etal{\emph{et al}\onedot}
\makeatother

\begin{document}
\pagestyle{headings}
\mainmatter
\def\ECCVSubNumber{16}  

\title{Identifying Auxiliary or Adversarial Tasks Using Necessary Condition Analysis for Adversarial Multi-Task Video Understanding} 


\titlerunning{Identifying Auxiliary or Adversarial Tasks Using Necessary Condition Analysis}
%
\author{Stephen Su\inst{1}\and
Samuel Kwong\inst{1} \and
Qingyu Zhao\inst{1} \and De-An Huang\inst{2} \and \\ Juan Carlos Niebles \inst{1} \and Ehsan Adeli \inst{1}}
\authorrunning{S. Su et al.}
%
\institute{Stanford University \\ \email{\{stephensu,samkwong,jniebles,eadeli\}@cs.stanford.edu}, \email{qingyuz@stanford.edu} \and
NVIDIA \\
\email{deahuang@nvidia.com}}
\maketitle

\begin{abstract}
There has been an increasing interest in multi-task learning for video understanding in recent years. In this work, we propose a generalized notion of multi-task learning by incorporating both auxiliary tasks that the model should perform well on and adversarial tasks that the model should not perform well on. We employ Necessary Condition Analysis (NCA) as a data-driven approach for deciding what category these tasks should fall in. Our novel proposed framework, Adversarial Multi-Task Neural Networks (\mtnet), penalizes adversarial tasks, determined by NCA to be scene recognition in the Holistic Video Understanding (HVU) dataset, to improve action recognition. This upends the common assumption that the model should always be encouraged to do well on all tasks in multi-task learning. Simultaneously, \mtnet still retains all the benefits of multi-task learning as a generalization of existing methods and uses object recognition as an auxiliary task to aid action recognition. We introduce two challenging Scene-Invariant test splits of HVU, where the model is evaluated on action-scene co-occurrences not encountered in training. We show that our approach improves accuracy by $\sim3$\% and encourages the model to attend to action features instead of correlation-biasing scene features.

\end{abstract}

\section{Introduction}

There has been an increasing interest to look beyond a single action label for video understanding~\cite{Damen2020Collection}. Even for action recognition, it is important for video models to look at multiple elements in a video, such as objects, scenes, people, and their interactions~\cite{diba2019HVU,bagautdinov2017social,ray2018scenes}. One prominent approach to achieve this is through \emph{multi-task learning (MTL)}, where video models are not only trained for recognizing actions (the primary task of interest) but also other auxiliary tasks such as recognizing objects and scenes. The assumption is that encouraging the model to do well on various auxiliary tasks improves the generality of the learned feature representation, which in turn improves the performance and generalization of action recognition.

\begin{figure}[t]
	\centering
  \begin{minipage}[c]{0.52\textwidth}    \includegraphics[width=0.95\linewidth]{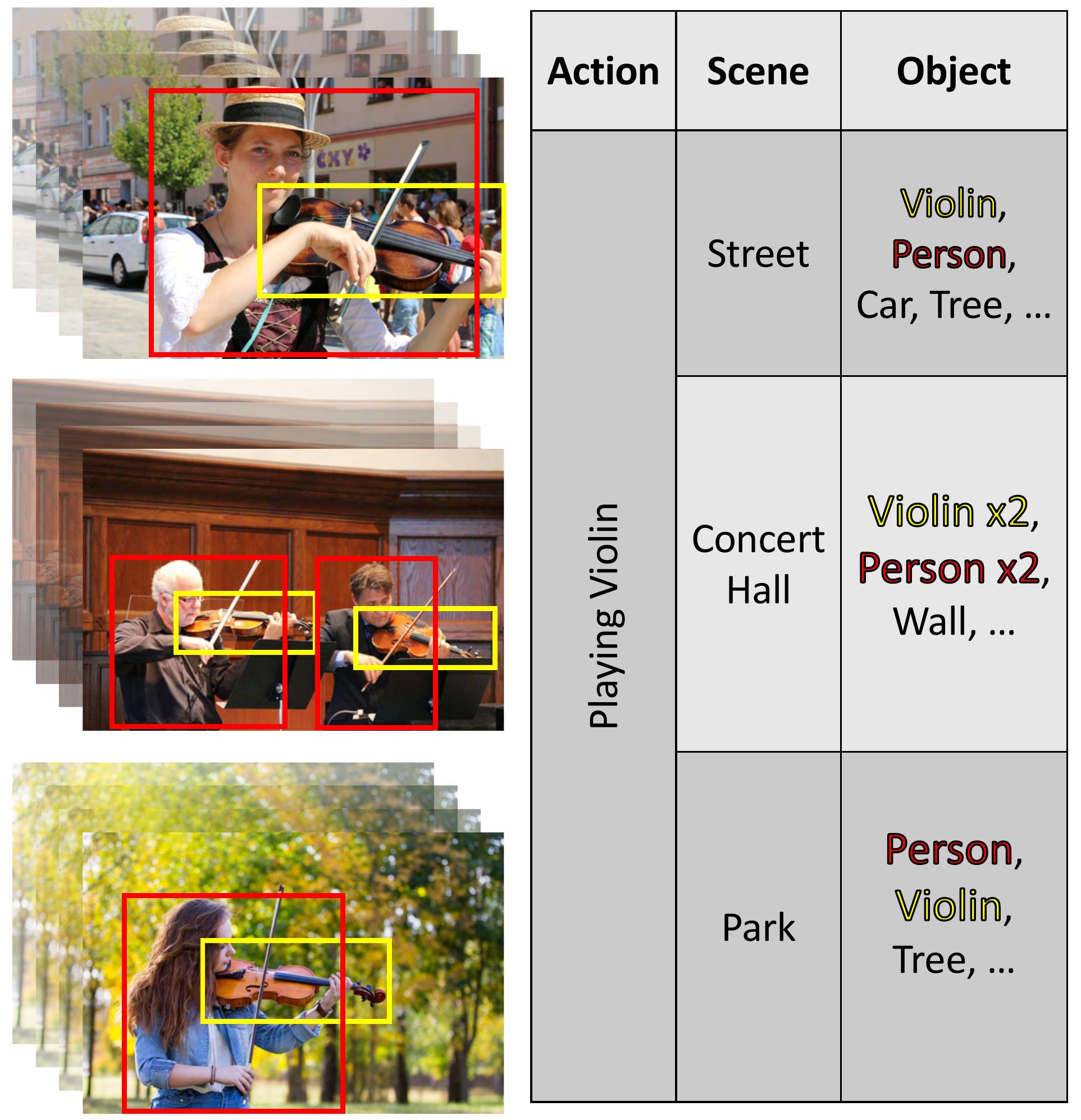} 
  \end{minipage}
  \begin{minipage}[c]{0.35\textwidth}	
  \caption{\small Action recognition models are often biased by the scene cues, as shown by the literature. This example illustrates that actions are not dependent on scenes but may be reliant on objects. Hence, object recognition as an auxiliary task can help action recognition. The same characteristic, however, may not hold for scenes. To force the model to not memorize scene cues, we propose multi-tasking while adversarially conditioning on scene.}
	\label{fig:fig1}
	\end{minipage}
\end{figure}

In this work, we show that this assumption for multi-task learning is not the best approach for video understanding. Instead of encouraging the model to do well on all auxiliary tasks, it could be better to encourage the model to \emph{not} do well on some of these tasks. For instance, in Fig.~\ref{fig:fig1}, while the violin object is crucial for properly performing the action of ``playing violin," the scene is not necessary for the action to occur. By encouraging the model to \emph{not} do well on recognizing the scene, we incentivize the model to not leverage the background bias when learning (i.e., recognizing an action by just looking at the background scene), which was shown to hurt the generalization of video models ~\cite{choi2019why,Huang2018WhatMA}.

Though it is possible to use action-scene correlations to try improving accuracy, we show empirically it does not consistently improve results and performs worse than \mtnet. A possible reason is that features learned while ignoring the scene are much more representative of an action. 	Further, unbalanced datasets suffer from selection bias \cite{winship_mare_1992}. To mitigate the effects of selection bias and encourage independent feature learning \cite{shen2018, adeli2019representation}, we use \mtnet and its negative branches. 

This inspires us to propose Adversarial Multi-Task Neural Networks (\mtnet). In contrast to existing works that only focus on beneficial auxiliary tasks, \mtnet further includes \emph{adversarial tasks} that the model should \emph{not} do well on. This is a broader generalization of multi-task learning that includes auxiliary or adversarial training on one or more tasks as a special case.

Instead of relying on intuition to determine what tasks are helpful or bias-inducing, we propose using Necessary Condition Analysis (NCA) \cite{dul2016necessary} as a data-driven method for differentiating between auxiliary and adversarial tasks. We choose the Holistic Video Understanding (HVU) dataset~\cite{diba2019HVU} to showcase our approach with \mtnet, as indicated in Fig.~\ref{fig:architecture}. All branches share a 3D feature extraction backbone, ensuring that the extracted features are predictive of the goal task(s) while not doing well on the undesired tasks. Our key observation for the HVU dataset is that while some scenes are highly correlated with the actions being performed in them, they are not necessary in order for the actions to be performed. For example, being in a casino is not necessary for playing blackjack, but the two have high correlation. On the other hand, holding a violin is necessary for playing violin. Therefore, we use scene recognition as an adversarial task and object detection as an auxiliary task for action recognition.

Our experiments show that \mtnet performs better on HVU compared to regular multi-task learning. Additionally, to evaluate the generalization of the video model, we propose the Scene-Invariant data splits of HVU, where videos with rare scenes are placed in the test set. In the new splits, we ensure that the testing set has videos with action:scene pairs that are never seen in the training phase. This extreme split evaluates if the trained models are dependent on the scenes or on the action cues. A higher relative performance on these splits translates to less bias for the action recognition models with respect to the background. We show that \mtnet more significantly outperforms existing multi-task learning approaches on this split. This indicates the importance of additionally leveraging adversarial tasks in the proposed \mtnet.

The main contributions of our work are as follows: (1) We introduce \mtnet, a general multi-tasking framework that leverages a shared feature extraction component across tasks but creates compromise on what tasks to do well on and not to do well on; (2) We propose a data-driven approach using NCA for identifying which tasks to positively multi-task on and which tasks to learn adversarially instead; (3) To evaluate our method, we provide the two Scene-Invariant splits of the recently introduced HVU dataset.

In summary, this work pushes the typical paradigm of multi-task learning---from jointly learning auxiliary tasks and discarding the rest---to instead identifying and learning both auxiliary and adversarial tasks in a data-driven approach. Our approach can be generalized to any dataset with multi-tasking attributes.

\section{Related Work}

In this section, we briefly review previous works that have elaborated on scene bias in video datasets and used approaches related to ours.

\vspace{1mm}
\noindent\textbf{Video Action Recognition.} Two-stream networks \cite {simonyan2014twostream} and 3D convolutional networks \cite{carreira2017quo,hara2017learning,tran2015learning} have improved human action recognition in recent years. 
One of the most important aspects of video action recognition is capturing the long-term temporal structure of the action being performed across video timesteps~\cite{wu2019long}. Current large-scale video datasets, however, provide many alternate cues over time that may influence the classifier's learned parameters~\cite{Damen2020Collection,goyal2017something,gu2018ava,ji2020action,ray2018scenes}. For example, the Holistic Video Understanding (HVU) dataset provides additional labels like scenes and objects~\cite{diba2019HVU}. Here, we propose a generalized MTL framework to better utilize these additional cues to learn video models that capture action-related information.

\vspace{1mm}
\noindent\textbf{Multi-Task Learning in Videos.} Most prior work on multi-task learning has focused on applications in natural language processing or single image learning. In videos, there is some preliminary work on better multi-task learning for videos~\cite{hong2018videomtl, kim2020mila,nguyen2019multitask,ouyang2019videomtl, zhu2017videomtl} and there has been increasing interest in the field. In contrast to existing works on multi-task learning, which assume that all the auxiliary tasks are beneficial to the action recognition performance, our proposed \mtnet further incorporates adversarial tasks. This further expands to space of possible tasks that could help multi-task learning. This is also different from selecting which tasks are useful for multi-task learning~\cite{standley2019tasks}, which still share the same assumption as existing works.

\vspace{1mm}
\noindent\textbf{Spatial Bias in Video Datasets.} The main difference between video and image data is temporal modeling, and, hence, temporal information in videos plays an important role in classifying the actions happening in a sequence~\cite{girdhar2019cater,sigurdsson2017actions}. Related works show that state-of-the-art 3D convolutional neural networks do not perform well on recognizing the action a subject is performing without explicit scene detail~\cite{chen2020deep,Huang2018WhatMA,weinzaepfel2019mimetics}. Huang \etal~\cite{Huang2018WhatMA} 
analyzes motion information in video datasets such as UCF101 and Kinetics. They found that around 40\% of classes in these datasets do not require motion to match the average class accuracy. 
Through their approach they concluded that motion was not important for classifying an action in a video and that these video datasets are not built in a way that makes temporal information relatively important for making action predictions. One way to address this is to create datasets that actually need temporal information for classification~\cite{sevilla2019only}. 
Li \etal~\cite{li2019repair} formulates bias minimization as an optimization problem, such that the dataset is resampled and example-level weights penalize examples that are easy for the classifier if built on a given feature representation. 
In this work, we show that multi-task learning is one alternative promising direction to mitigate bias of irrelevant spatial scene features for action recognition. 

\vspace{1mm}
\noindent\textbf{Adversarial Training in Videos.} Previous works have taken an approach of adversarial training for domain adaptation and bias mitigation ~\cite{adeli2019representation, akuzawa2020adversarial, choi2019why, elazar-goldberg-2018-adversarial, ganin2016domain, xie2018controllable, zhang2018mitigating}. The purpose of adversarial training is to encourage the model to learn features indicative of the original domain while also being invariant to the change in domains. With adversarial training, learned features may contribute to not only unbiased predictions but also better prediction performance.

In particular, the method in ~\cite{choi2019why} uses adversarial training for video action recognition. This work proposes a debiasing approach by introducing a scene adversarial loss in the pre-training step---with scene labels on video samples attained via a pre-trained scene classifier---and a human mask confusion loss by masking out humans in the video. In contrast, our work 1) does not assume scenes should be trained adversarially in all datasets and instead uses NCA to determine task specification, 2) proposes a more general framework of multi-task learning that includes auxiliary tasks such as object/attribute recognition to help the primary task, and 3) maintains adversarial training throughout the training step instead of only the pre-training step so as to ensure biases are not present after training. We apply our method on the less noisy HVU dataset and verify the feature extractor is being affected by the heads (see supplement). 


\section{Dataset and Necessary Condition Analysis}

\subsection{HVU Dataset}
The Holistic Video Understanding (HVU) dataset \cite{diba2019HVU} contains 572k video clips of humans performing different actions. We use action labels where each video corresponds to one primary action label. Importantly, beyond action labels, the videos in the dataset contain additional labels in one of five categories: ``Scene," ``Object," ``Event," ``Attribute," and ``Concept." In this work, we use the scene and object labels. We experiment on three different splits of the HVU dataset. One is the original split defined in the paper. It contains 572k videos, 882 action labels, 282 scene labels, and 1917 object labels. There are 476k videos in the training set and 31k videos in the validation set. The two other Scene-Invariant splits are defined in Section \ref{Scene-Invariant splits}.

\subsection{Necessary Condition Analysis (NCA)}
\label{NCA}
In some scenarios, an object or scene classifier can be used to aid action recognition. For example, the presence of an elliptical trainer is a necessary condition for the action of elliptical workout, so an object classifier, in this case, should be formulated as a multi-task learning objective to help the classification of the action. However, in some other scenarios, the presence of the object or scene highly correlates with but does not necessarily induce the action. For example, the action of playing blackjack does not depend on the location but has a high chance of being in a casino (high correlation). In this case, the tendency to recognize the casino scene can introduce bias to action recognition. We observe that the key difference between the two aforementioned scenarios is whether the object or scene is a necessary condition for the action. In this study, we resort to Necessary Condition Analysis (NCA) \cite{dul2016necessary} to explore the relationship between object, scene, and action labels.

\begin{figure}[t]
	\centering
	\vspace{-7pt}
	    \includegraphics[width=0.2\linewidth,trim=155 122 545 122, clip]{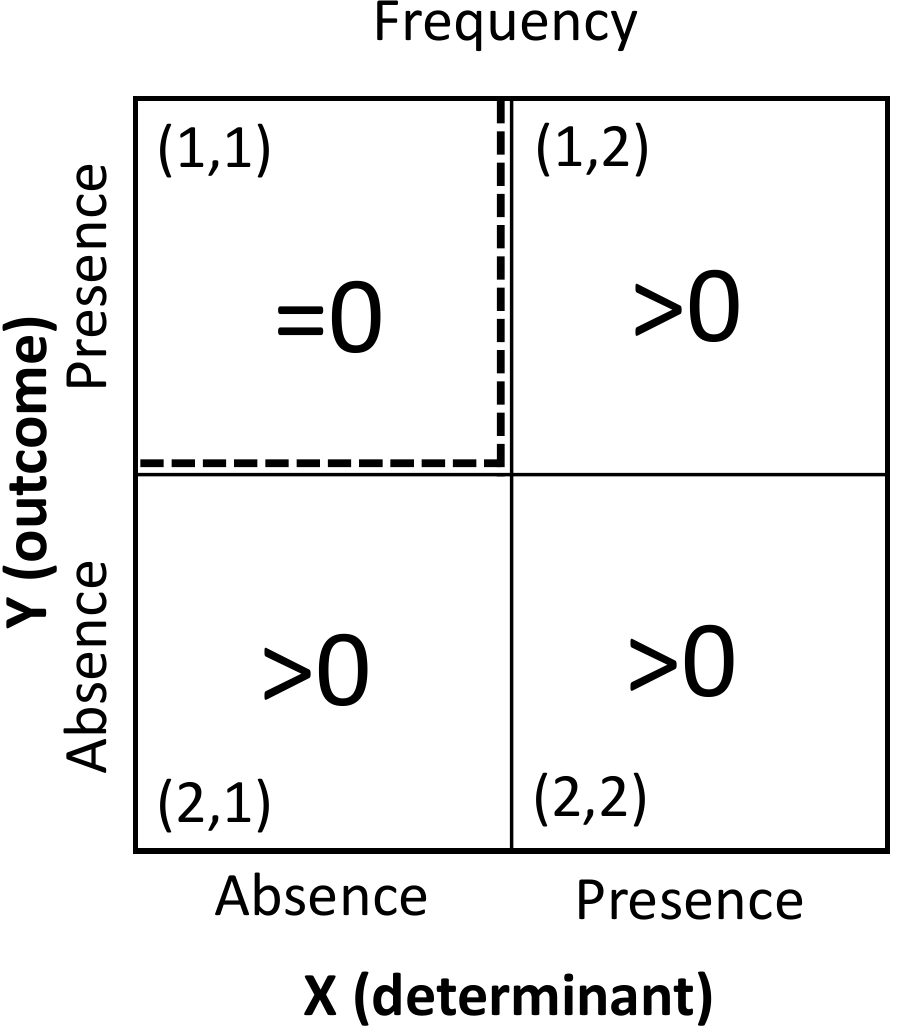}
	    \includegraphics[width=0.2\linewidth,trim=420 122 280 122, clip]{images/NCA_qz.pdf}
	    \includegraphics[width=0.2\linewidth,trim=680 120 21 122, clip]{images/NCA_qz.pdf}
	~\includegraphics[width=0.37\linewidth,trim=0 5 0 5, clip]{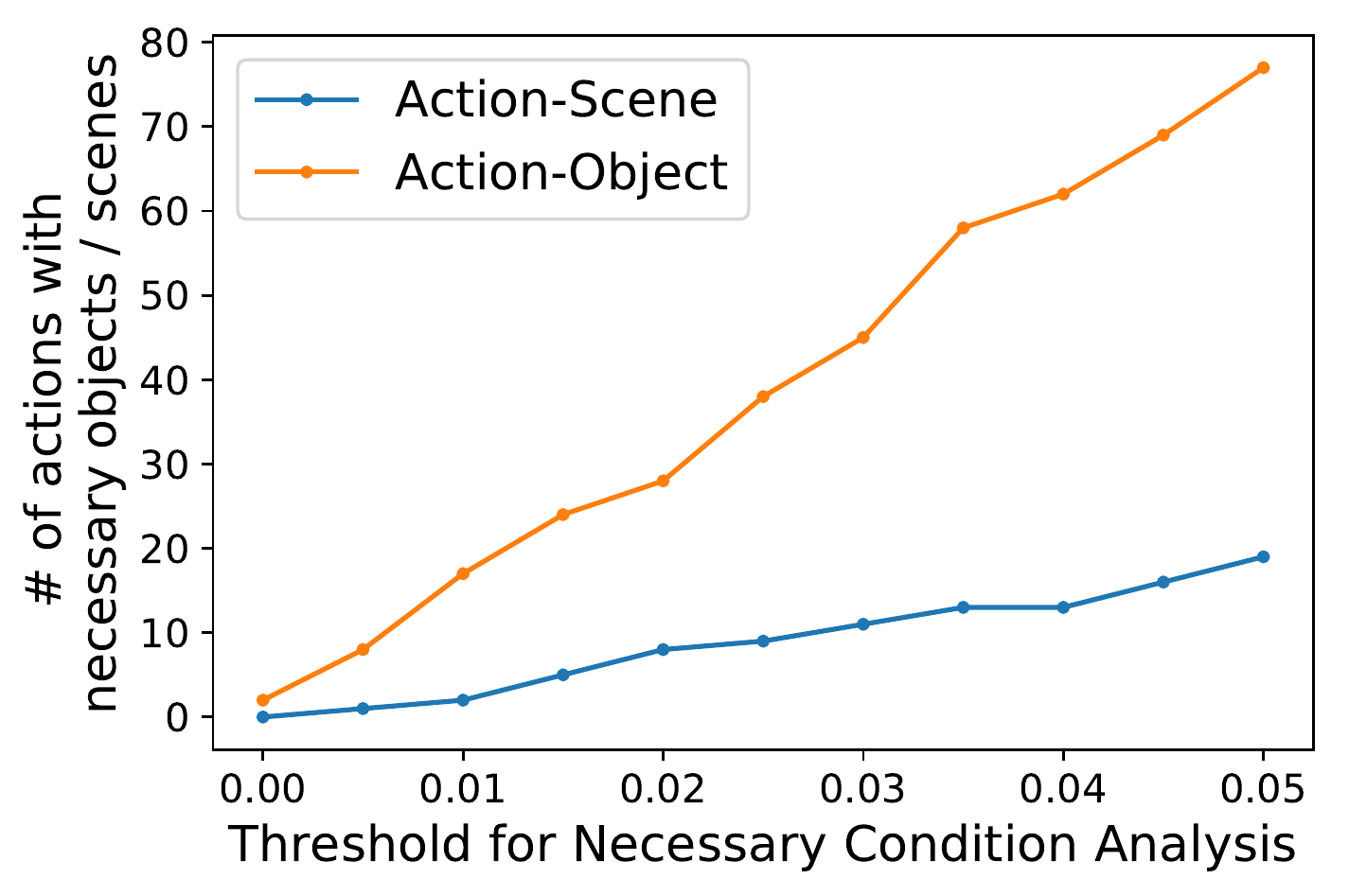}
	\vspace{-15pt}
	\caption{\small (a) The dichotomous necessary condition with possible combinatorial status of X and Y recorded in a $2 \times 2$ contingency matrix. (b, c) Average normalized contingency for action-object and action-scene pairs. (d) Number of actions with necessary objects and scenes for different NCA threshold.}
	\vspace{-10pt}
	\label{fig:NCA}
\end{figure}

\vspace{1mm}
\noindent\textbf{Dichotomous Necessary Condition.} X (object or scene) is said to be a necessary determinant of outcome Y (action) when X must be present for achieving Y (but its presence is not sufficient to obtain Y). Without the necessary condition, there is guaranteed absence of Y, which cannot be compensated by other determinants of the outcome. This logic is fundamentally different from the traditional correlation or regression analysis. In practice, the dichotomous NCA can be examined by a $2 \times 2$ contingency matrix, which counts the frequency of the combinatorial status of X and Y for their dichotomous logic (Fig. \ref{fig:NCA}a). X is a necessary condition of Y if and only if all observations are located below the dashed line (known as the ceiling line). In other words, one should not observe the outcome Y in the absence of X. 

We now analyze the necessity of objects and scenes for action recognition using the above logic. For each action, we first select the object that co-occurs most frequently and record the $2 \times 2$ contingency matrix for that action-object pair. We then normalize the matrix based on the frequency of co-occurring (i.e., the number in entry (1,2) of Fig. \ref{fig:NCA}b) and compute the average matrix across all actions. Note, we do not explicitly measure the frequency of both the action and object being absent (entry (2,1)  of Fig. \ref{fig:NCA}b) because that frequency can be arbitrarily inflated by introducing irrelevant items into the dataset. Finally, we repeat this procedure of computing average contingency matrix for the action-scene pairs (Fig. \ref{fig:NCA}c). We observe that on average, objects are more likely to become necessary conditions for action recognition as the number above the ceiling line is closer to 0. In this case, object recognition is more likely to contribute to action recognition in a MTL setting. On the other side, the scene labels are less likely to be necessary for the actions, so learning cues in the video related to the scenes are likely to bias the action classifier. Lastly, we compute the number of actions with necessary objects or scenes by examining the contingency matrix of each action-object or action-scene pair. Since real-world datasets are subject to noisy observations and inconclusive labeling, we relax the constraint that entry (1,1) of the contingency matrix has to be exactly 0 for necessary conditions. Instead, the necessary condition holds as long as entry (1,1) is less than a threshold. Fig. \ref{fig:NCA}(d) shows that for different thresholds the number of actions with necessary objects is always greater than the number of actions with necessary scenes.

\subsection{Human Study/Necessity Annotation}
\label{Human Study}
To motivate our approach and verify the results of Section \ref{NCA}, we conduct a human study to create a ``necessity annotation" that determines the relationships between the actions, scenes, and objects present in the HVU dataset. Specifically, we aim to determine which actions require or do not require a particular scene or object. We extract all occurrences of actions and their corresponding scenes and objects. There are a total of 31k unique action-scene pairs and 145k unique action-object pairs present in the HVU dataset.

During the study, participants are instructed on the difference between a ``necessary" and ``sufficient" condition and examples are provided to verify understanding.   Participants are then presented with the name of an action and  the name of a scene or object.  For every action-scene pair, we ask participants to answer the question: \emph{Is the scene necessary for the action to be correctly performed?} Similarly, for every action-object pair, we ask participants: \emph{Is the object necessary for the action to be correctly performed?}

An example of an action-scene pair that participants agree the scene is not necessary for the action to be performed is \emph{breakdancing} (action) and \emph{stage} (scene). On the other hand, a pair where the scene is necessary for the action to be correctly performed is \emph{ice swimming} (action) and \emph{body of water} (scene). For action-object, a pair that participants agree the object is not necessary for the action to be correctly performed is \emph{climbing tree} (action) and  \emph{shoe} (object). In contrast, participants agree the object is necessary for the pair \emph{playing violin} (action) and \emph{violin} (object).

We find that \textbf{8.02\%} of actions require a scene in order to be correctly performed, with 0.0928\% variance between three participants. Additionally, \textbf{98.9\%} of actions require an object in order to be correctly performed, with 0.00429\% variance between participants. From this study, we observe that scenes are mostly unnecessary for actions to be properly performed while objects are much more necessary for actions to be properly performed.

\subsection{Scene-Invariant Splits of the HVU Dataset}
\label{Scene-Invariant splits}

Recall from Sections \ref{NCA} and \ref{Human Study} that an action can have highly correlative features such as a particular scene associated with it. However, these features may not be necessary or relevant for the action to be performed. Learning these actions becomes difficult as it is hard for models to disentangle relevant and irrelevant features for an action. 

In fact, deep learning models will typically first learn lower-level spatial features rather than more complex features. This is particularly true in the context of video action classification as the most relevant features of a video are in the complex temporal stream, but models can rely on simpler spatial features to achieve similar accuracies if there is high correlation with those spatial features. An action that has a highly correlative spatial feature, for example, is \emph{playing soccer} as most videos of soccer will have grass present. A naive model may not be learning the actual human playing soccer, but rather relying on features for grass and associating that with playing soccer.

In this work, we hope to specifically test models for learning an action's most relevant features and reduce reliance on irrelevant features like certain spatial features. As a result, we look to assess a model's robustness to spatial bias. To meet this goal, we introduce two \textbf{Scene-Invariant} training and validation splits of the HVU dataset which serve to test a model's reliance on scene while performing action recognition.

To create the Scene-Invariant 1 split, we filter the HVU dataset to contain only videos that have  a scene label and an action category that includes at least 124 videos. For videos with multiple scene labels, only one label is selected for downstream analyses. We choose this scene label by finding the scene that occurs most frequently within a video's action class and breaking ties using global scene popularity among all action classes. This way, if a video's chosen scene is classified as ``rare," then we are guaranteed that every other scene in the video, if it exists, will be even more ``rare" than the chosen scene. Next, we group videos with the same action and scene together in action:scene pairs and take the rarest action:scene pairs to use in the validation set. The remaining videos are moved to the training set. The end result is a training set consisting of the most common action:scene pairs and a validation set with the rarest action:scene pairs. Note that this task is particularly difficult because the action:scene pairs tested in the validation set are never seen in the training set. 

To create the Scene-Invariant 2 split, we again refine the HVU dataset to only videos that contain scene labels and action categories that have at least 124 videos.  For videos with multiple scene labels, we use the scene label that is most rare within a video's action class and break ties using global scene popularity among all action classes.  This way, if a video's chosen scene is classified as ``common," then we are guaranteed that every other scene in the video, if it exists, will be even more ``common" than the chosen scene. Next, we group videos with the same action and scene together in action:scene pairs and take the most common action:scene pairs to use in the training set.  The remaining videos are moved to the validation set. Note that this task is particularly difficult because the model only ever sees a few scenes associated with each action and that action:scene pairs tested in the validation set are never seen during training. 

To summarize, the Scene-Invariant 1 split has the most rare action:scene pairs placed in the validation set while the Scene-Invariant 2 split has the most common action:scene pairs placed in the training set.  In both Scene-Invariant splits, we prevent a model's ability to rely on irrelevant scene features by creating a validation split with actions in new scene contexts that do not show up in the training set.  A better performance on the Scene-Invariant splits translates to a model learning the action more effectively by relying on action features.

\subsection{Evaluation Metric for Scene Bias}
For analysing the correlation of learned features and the scene labels, we use a well-studied statistical dependence of two vectors (features vector and the one-hot scene label vector) based on squared distance correlation ($dcorr^2$) \cite{szekely2007measuring}. Classical tests such as the Pearson correlation only measure linear dependence. In contrast, $dcorr^2=0$ if and only if there is complete statistical independence between two random variables. We calculate $dcorr^2$ between the extracted features and its associated one-hot scene label vector.  Intuitively, a high $dcorr^2$ indicates it is more possible to predict scene labels from extracted features, while a low $dcorr^2$ indicates it is not possible to predict the scene labels as there is no scene information contained in the extracted features.  

\section{Method}

\begin{figure}[t]
	\centering
	\includegraphics[width=0.87\textwidth]{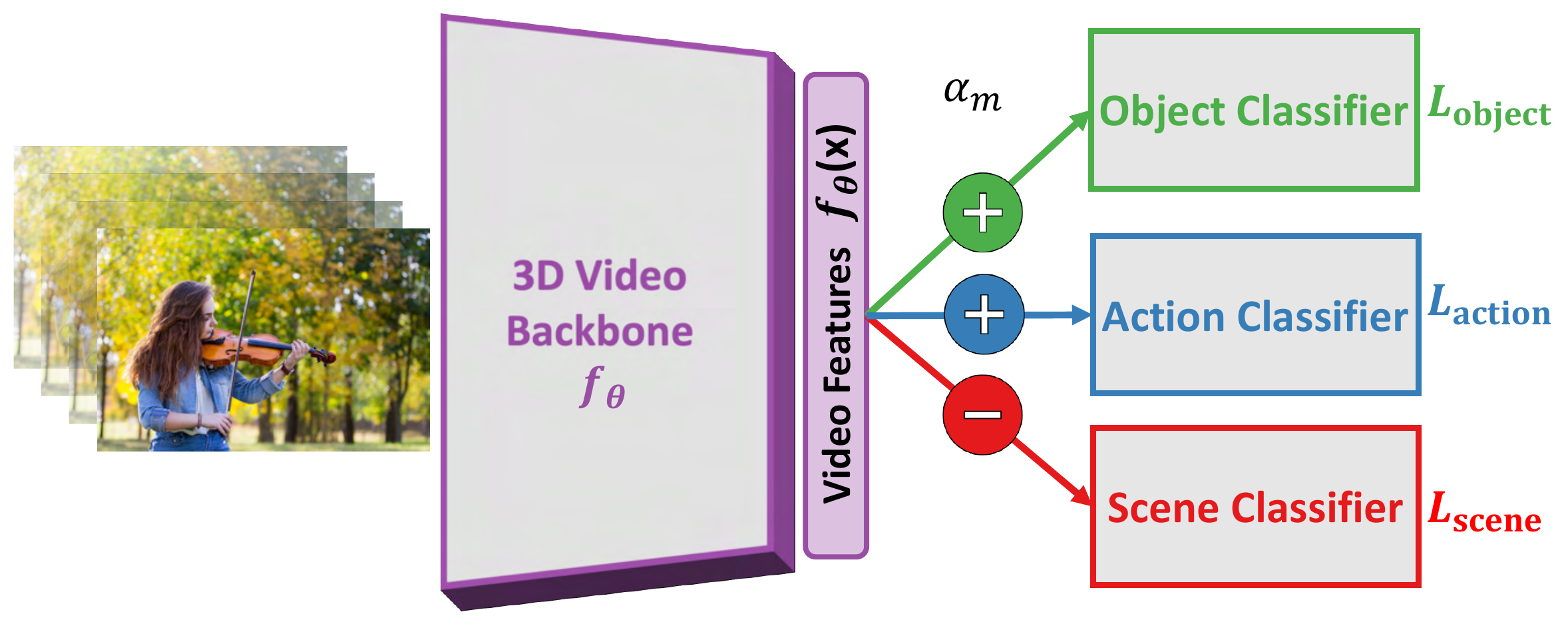}
	\caption{\small Overview of \mtnet: For the actions in the HVU dataset, NCA determines that object recognition should be an auxiliary task while scene recognition  should be an adversarial task.  \mtnet learns action/object features while ignoring scene features.}
	\label{fig:architecture}
\end{figure}

We introduce a supervised-learning algorithm that seeks to learn a single feature representation, $f_\theta(x)$, which is in turn used as a feature representation for downstream classification tasks (calculated using the feature extractor $f_\theta$). 

Let $x$ denote the input videos and $y$ denote the corresponding labels.  For simplicity, the primary task is only associated with single labels, \ie, every video is associated with one label, but the general approach is compatible in a multilabel setting for the primary task. Furthermore, in this work we focus on videos to work on the scene bias problem, but the input $x$ could be an image or video clip. The objective for the primary classification task, with $K$ classes, and its corresponding classifier, $g$, will be a cross entropy loss denoted as 
\begin{align}
	L\left(x, y | f_\theta(x), g \right) & =- \sum_{k=1}^{K} \mathds{1}[y=k] \log p_{k}          
	                                                            = L_{action}\left(x, y | f_\theta(x), g \right) 
\end{align}
where $p_k$ is the normalized softmax output of the classifier. As this classification loss is on the primary task, in this case action recognition, we can refer to this loss as $L_{action}\left(x, y | f_\theta(x), g \right)$.

Now, we define the objectives associated with the other tasks. We denote $g_t$ as the classifier function associated with task $t$. Since these tasks are in a multi-label setting, we use the binary cross entropy loss
\begin{equation}
    \begin{aligned}
        L_{t}(x, y_t, g_t | f_\theta(x)) = - \sum ^{K_t}_{k=1} \left((y_{t,k}) \log(p_k)  + (1-y_{t,k}) \log(1-p_k) \right)  
    \end{aligned}
\end{equation}
where $K_t$ denotes the number of classes in task $t$ and $y_{t,k}$ denotes the binary label associated with class $k$ of task $t$. We will refer to the binary cross entropy loss associated with task $t$ as $L_t(x, y_t, g_t | f_\theta(x))$.

Now, we combine the individual losses to optimize the joint multi-task learning objective below:
\begin{equation}
    L_{total} = L_{action} + 
\sum_{t=1}^T \alpha_t L_t
\end{equation}
where $\alpha_t$ determines the contribution of each of the losses associated with the $T$ additional tasks. Importantly, the sign of $\alpha_t$ can be either positive or negative. We refer to this generalized framework for multi-task learning as \mtnet. 

In the experiments, 3D-ResNet \cite{hara2017learning} will act as the feature extractor $f_\theta$ and $L_{action}$ refers to the loss of the primary task, action classification. $L_t$ encompasses $L_{scene}$, the scene loss, and $L_{object}$, the object loss.  Throughout the paper, $\lambda$ will refer to the contribution of $L_{\text{scene}}$ and $\gamma$ will refer to the contribution of $L_{\text{object}}$  ($\lambda < 0$ and $\gamma > 0$). In the three branch version of the model, we can interpret minimizing this objective function as encouraging a model to learn object features to be used by the object branch to successfully classify objects, and to unlearn scene features so that the scene branch does not have the capability to classify scenes. Finally, $dcorr^2$ is calculated between the extracted features $f_\theta(x)$ and the associated  one-hot encoded scene labels.


\setlength{\tabcolsep}{4pt}
\begin{table}[t]
    \begin{center}
        \caption{\small Action accuracy on the Modified 20BN-Jester Dataset.}
        \label{table:sife_jester_results}
        \begin{tabular}{lll}
            \hline\noalign{\smallskip}
            Method & Train Acc. & Val Acc.\\
            \noalign{\smallskip}
            \hline
            \noalign{\smallskip}
            Baseline I3D & 92.74 & 87.97\\
            {\mtnet (Ours)} & {\bf 94.60} & {\bf 91.58}\\
            \hline
        \end{tabular}
    \end{center}
\end{table}
\setlength{\tabcolsep}{1.4pt}

\begin{figure}[t]
	\centering
	\includegraphics[trim=30 80 30 80,clip,width=\linewidth]{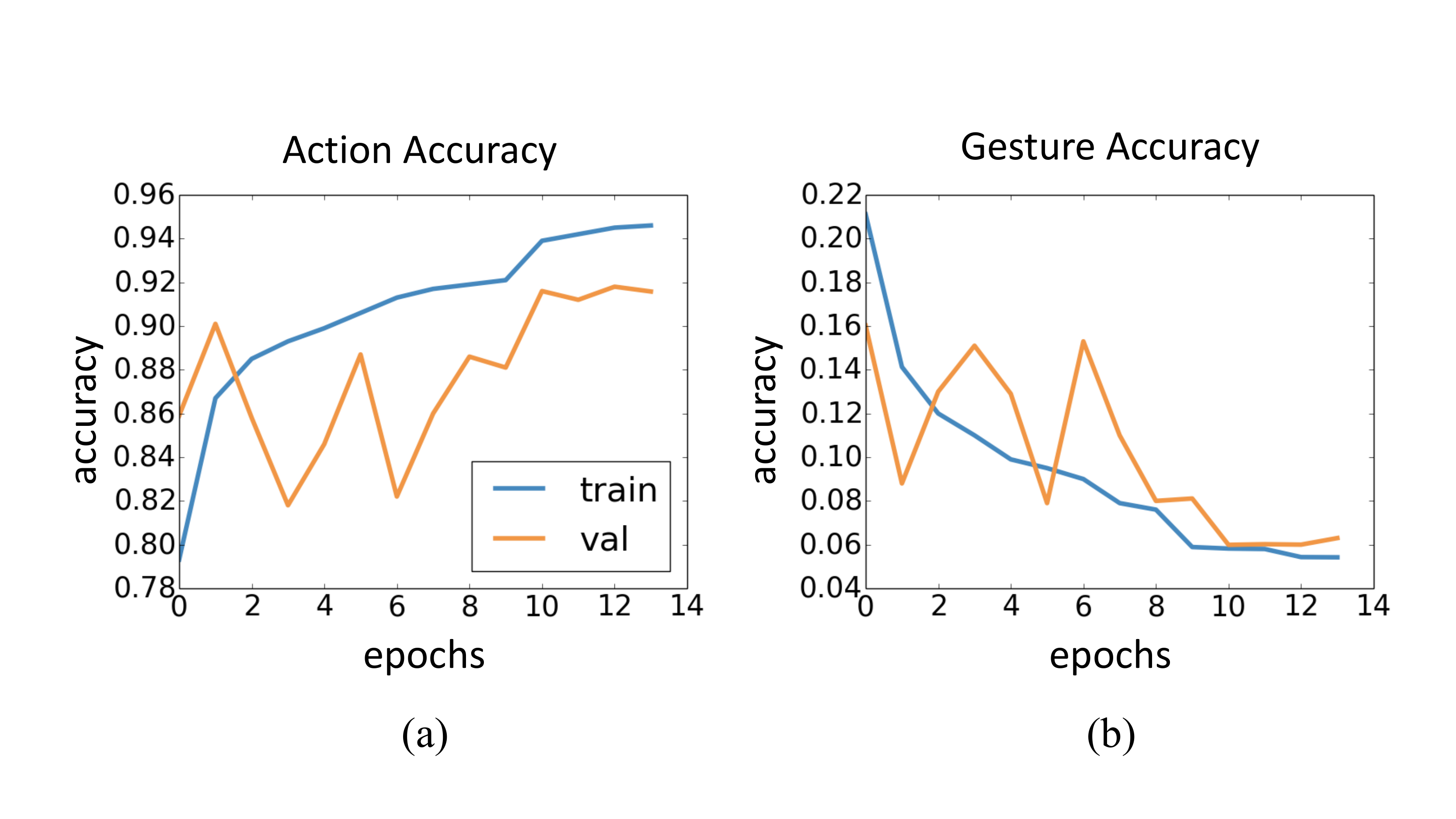}
	\vspace{-20pt}
	\caption{\small \mtnet accuracy: (left) action \& (right) gesture (adversarial component) as a function of training iterations. As action recognition performance increases, gesture recognition performance decreases, an effect of the adversarial conditioning.}
	\label{fig:jester_curves}
\end{figure}


\section{Experiments}

In this section, we outline our experimentation details and report results on a synthetic setup to evaluate the argument that adversarial loss can remove the effects of bias cues in videos. Then, we report results of \mtnet on the HVU dataset and compare them with several baselines quantitatively and qualitatively. 

\subsection{Implementation Details}
\noindent\textbf{Synthetic Experiments.} We use a Two-Stream Inflated 3D ConvNet (I3D) \cite{carreira2017quo} pretrained on the Kinetics Human Action Video Dataset \cite{kay2017kinetics} as our baseline and feature extractor. The model is fine-tuned on our synthetic data. We use an Adam optimizer with an initial learning rate of $1 \times 10^{-2}$ and learning rate decay by a factor of 10 after every 10\textsuperscript{th} epoch. We then use \mtnet to decouple synthetic bias cues from actions cues and show our results in Section \ref{Synthetic Exp}.

\begin{figure}[t]
    \centering
    \setlength{\tabcolsep}{0pt}
    \begin{tabular}{cc}
        \includegraphics[trim=5 30 10 20,clip,width=0.35\textwidth]{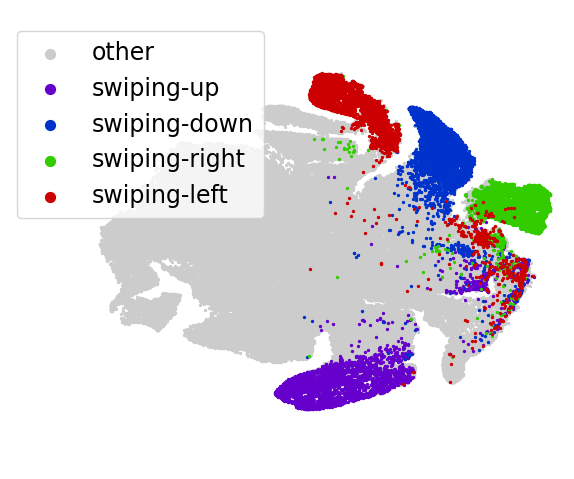}
        &
        \includegraphics[trim=5 30 10 20,clip,width=0.35\textwidth]{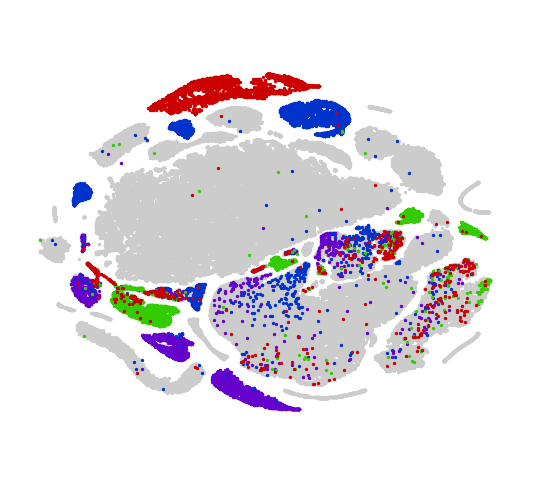} \\
        (a) Baseline & (b) \mtnet
    \\
    \end{tabular}
    \caption{\small Feature embedding space of the baseline and \mtnet on the Modified 20BN-Jester Dataset, using I3D backbone. \mtnet extracts features in the embedding space with less correlation bias.}
    \label{fig:tsne_i3d_action_jester}
\end{figure}

\setlength{\tabcolsep}{4pt}
\begin{table*}[h]
	\begin{center}
		\caption{\small Results of the HVU dataset with Default and Scene-Invariant Splits. In the Scene-Invariant setting, accuracy is improved. We report the top-1 accuracy (Acc; higher better) and the distance correlation ($dcorr^2$; lower better) of learned features and the scene labels. `-' means not applicable. $\oplus$ and $\ominus$ refer to non-adversarial and adversarial classifications, respectively. A: Action, S: Scene, O: Object. \vspace{-15pt}}
		\label{table:mtnet_results}
		\setlength{\tabcolsep}{2pt}
        \resizebox{\textwidth}{!}{%
        \begin{tabular}{llcccccccccccc}
			\hline\noalign{\smallskip}
			\multirow{2}{*}{\textbf{Method}} & \multirow{2}{*}{\textbf{Backbone}} & \multicolumn{3}{c}{\textbf{Task}} & ~ & 
			\multicolumn{2}{c}{\cellcolor{gray!25}\textbf{Split: Default}} & \cellcolor{gray!25}~ & 
			\multicolumn{2}{c}{\cellcolor{gray!25}\textbf{Scene-Invariant 1}} &\cellcolor{gray!25} ~ & 
			\multicolumn{2}{c}{\cellcolor{gray!25}\textbf{Scene-Invariant 2}} \\
			\cline{3-5}\cline{7-8}\cline{10-11}\cline{13-14}
			~                                                   & ~            & A   & S     & O   & ~ & Acc        & $dcorr^2$ & ~ & Acc            & $dcorr^2$            & ~            & Acc      & $dcorr^2$    \\
			\noalign{\smallskip}
			\hline
			\noalign{\smallskip}
			Baseline                                            & 3D-ResNet-18 & $\oplus$ & -         & -        & ~ & 43.40        &   0.303    & ~ & 30.77          &   0.338               &              & 39.87        & 0.409        \\
			Action-Scene MTL                                    & 3D-ResNet-18 & $\oplus$ & $\oplus$  & -        & ~ & 43.73        & 0.311      &   & 30.12          &    0.351              &              & 39.63        & 0.422        \\
			\mtnet (Ours)                                       & 3D-ResNet-18 & $\oplus$ & $\ominus$ & $\oplus$ & ~ & \textbf{44.06}  &  \textbf{0.281} &   & \textbf{33.86} &       \textbf{0.265}  &  & \textbf{43.94}  & \textbf{0.388}        \\
			\hline
			Baseline                                            & 3D-ResNet-50 & $\oplus$ & -         & -        &   & 52.63      &  0.333   &   & 36.86          &   0.343               &              & 45.12        & 0.423        \\
			Pretrain-scene Adv. \cite{choi2019why}        & 3D-ResNet-50 & -        & -         & -        &   & 51.35      &  0.321     &   & 37.18          &   0.339               &              & 45.74        & 0.395        \\
			Action-Scene MTL                                    & 3D-ResNet-50 & $\oplus$ & $\oplus$  & -        &   & 52.25       &    0.344   &   & 37.20      &     0.352             &              & 44.73        & 0.442        \\
			{\mtnet w\textbackslash o object}            & 3D-ResNet-50 & $\oplus$ & $\ominus$ & -        &   & {52.12} &  \textbf{0.311} &   & {38.52} &   \textbf{0.287}               &              & 47.12        & \textbf{0.372}        \\
			{\mtnet (Ours)} & 3D-ResNet-50 & $\oplus$ & $\ominus$ & $\oplus$ &   & \textbf{53.17} & {0.346} &   & \textbf{39.53} &    {0.321}    &              & \textbf{48.44}        & 0.379        \\
			\hline
		\end{tabular}
		}
	\end{center}
\end{table*}
\setlength{\tabcolsep}{1.4pt}

\vspace{1mm}
\noindent\textbf{HVU Experiments.} For our HVU experiment, we choose the 3D-ResNet architecture as our feature extractor and run experiments on 3D-ResNet-18 and 3D-ResNet-50. Models are pretrained on the Kinetics-700 dataset \cite{carreira2019kinetics} before running on the HVU dataset. We train with stochastic gradient descent (SGD) and initialize our learning rate as $1 \times 10^{-2}$, dividing by a factor of 10 when validation action accuracy saturates. The activation after the fourth ResNet block is used as the feature representation $f_\theta(x)$, the output of the feature extractor $f_\theta$. In the action branch, we apply a fully connected layer onto the extracted features to attain the final class scores. In the scene and object branches, we use two sets of linear, batchnorm, ReLU layers before a final fully connected layer. The hidden layers have 2048 units. We choose weights of the scene and object branches, $\lambda$ and $\gamma$, via cross-validation for each respective model and split. We train our network until validation accuracy saturates.  

We also evaluate a separate model from \cite{choi2019why} that was originally focused on transfer learning.  The model undergoes adversarial training on one dataset and the weights, excluding the adversarial head, are transferred to a vanilla model for a classification task on a separate dataset. For a fair comparison, we adapt this implementation using adversarial pretraining  on the HVU dataset and fine-tuning on the same dataset (denoted by ``Pretrain-scene Adv." \cite{choi2019why} in Table \ref{table:mtnet_results}).

\subsection{Synthetic Data Experiments}
\label{Synthetic Exp}
To evaluate the efficacy of our proposed adversarial component, we first set up a simple experiment with a synthetic dataset with videos borrowed from 20BN-Jester \cite{jester}.

\vspace{1mm}
\noindent\textbf{Modified 20BN-Jester.} The 20BN-Jester Dataset contains videos of humans performing pre-defined hand gestures. Many of these hand gestures are mirrored (\eg, swiping left/right, swiping up/down). As a result, action classes may be distinct in terms of direction (\eg, left,  right) but similar when considering the overall gesture category (\eg, swiping). We create 5 action classes of \emph{swiping-left}, \emph{swiping-right}, \emph{swiping-up}, \emph{swiping-down}, \emph{others} and 2 gesture classes of \emph{swiping} and \emph{other} and call this synthetic dataset the Modified 20BN-Jester Dataset. Our goal is to decouple action cues from gesture cues.

\vspace{1mm}
\noindent\textbf{Results.} Fig.~\ref{fig:jester_curves} shows that while action recognition improves after each epoch, recognizing the overall gesture (common across all actions) worsens as intended.

As shown in Table \ref{table:sife_jester_results}, \mtnet outperforms the I3D baseline on the Modified 20BN-Jester Dataset after 12 epochs by 1.86\% in training and 3.61\% in validation. This boost in performance serves as a proof of concept that mitigating correlation bias via our adversarial component can lead to an improvement in action prediction.

\vspace{1mm}
\noindent\textbf{Visualizing Action Feature Space.} Further, we visualize the extracted features from our baseline and \mtnet using two-dimensional t-SNE plots in Fig. \ref{fig:tsne_i3d_action_jester}. We distinguish all different swiping directions by separate colors. Fig. \ref{fig:tsne_i3d_action_jester}(a) shows that while video samples are generally clustered together by their action directions, all videos of a swiping gesture are clustered together by that gesture as well. Fig.~\ref{fig:tsne_i3d_action_jester}(b), however, demonstrates that our model learns features from video samples that are not biased by the swiping gesture, as is represented by the wider distribution of samples in the feature space. While \mtnet can better classify the fine actions (Table \ref{table:sife_jester_results}), the feature embedding space is not driven by the overall gesture of ``swiping."

\begin{figure*}[t]
	\centering
	\includegraphics[trim=60 10 60 10 ,clip,width=\linewidth]{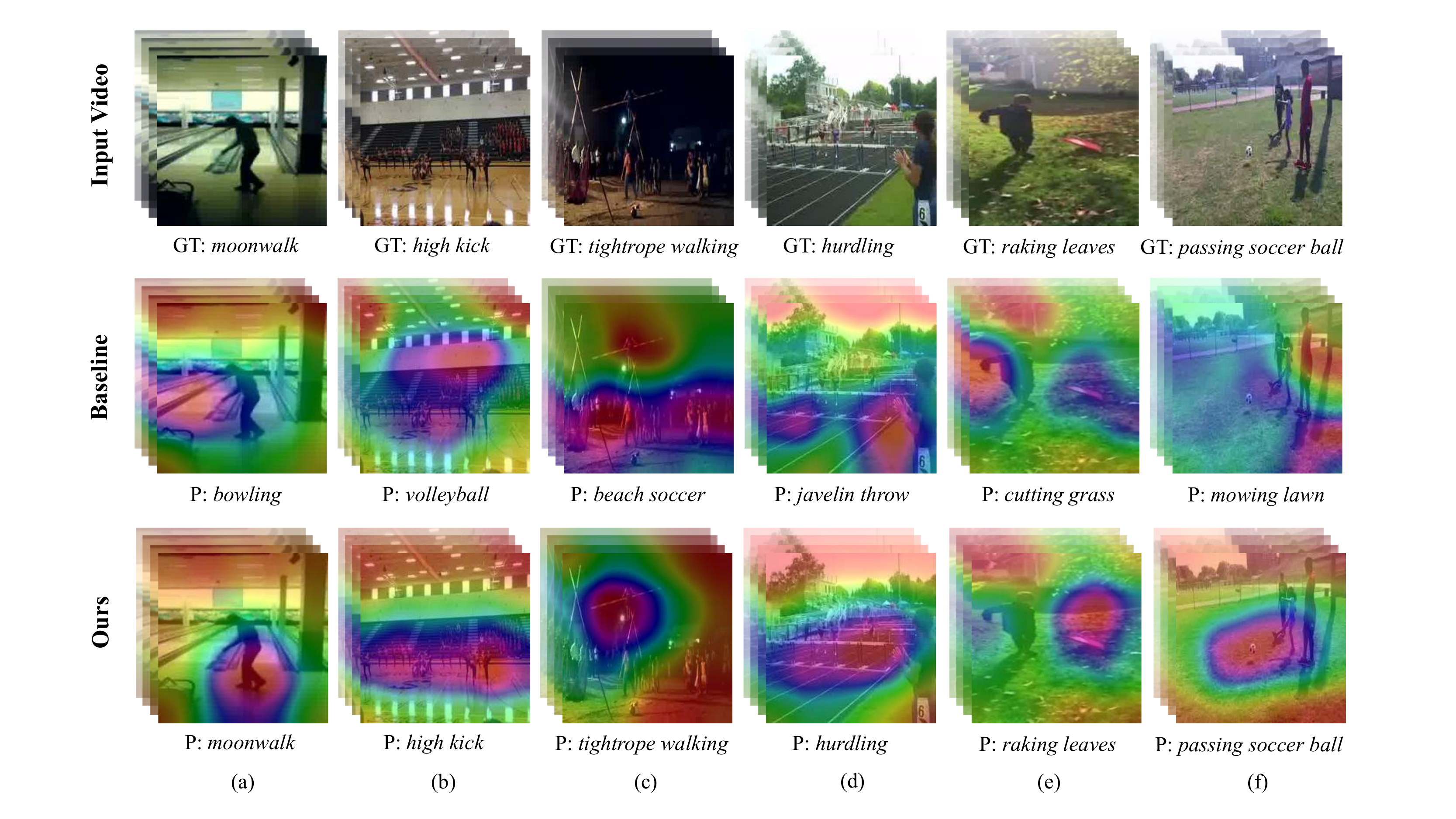}
	\vspace{-20pt}
	\caption{\small Qualitative results on six videos from HVU. For each video, we list the action ground truth (GT) and collect the action prediction (P) and saliency map from the 3D-ResNet-50 baseline and compare to those of \mtnet (Ours). Red regions represent higher activation scores and greater attention from the trained model.} 
	\label{fig:saliency}
\end{figure*}

\subsection{HVU Dataset Experiments}

\noindent\textbf{Results.} 
Table \ref{table:mtnet_results} shows the performance of \mtnet applied on the three splits of the HVU dataset. For the default dataset split, the data distributions of the training and validation set are similar so corresponding actions and scenes are still correlated in the validation. Therefore, a model using action-scene correlations can still work well. Conveniently, \mtnet still performs slightly better than baseline despite being unable to rely on the scene information.

In the Scene-Invariant splits, \mtnet consistently outperforms the baseline. This validates the idea that \mtnet actively incorporates relevant action features to use in the prediction, while the baseline model relies on irrelevant spatial features.  \mtnet is therefore more robust to changes in irrelevant spatial features. Furthermore, the additional object branch in \mtnet acts synergistically with the scene branch and consistently improves outcomes to become more accurate than the other baselines including the Pretrain-scene Adv. model.  The likely explanation for the performance of \mtnet over the Pretrain-scene Adv. model is that the Pretrain-scene Adv. model is only adversarially trained during the pretraining step. When it is fine-tuned, the adversarial step is removed and the model can relearn its original biases, decreasing its robustness to irrelevant scenes.  We also observe that the weaker Resnet-18 models  see larger improvements with \mtnet as weaker models are more prone to overfitting to the bias. Our adversarial conditioning regularizes the model and forces it to avoid the easy cues.

Of note, \mtnet performs the best on the Scene-Invariant 2 split.  This is distinctive as the Scene-Invariant 2 is the most comprehensive assessment of robustness to irrelevant spatial features. In this split, actions in the training set are only associated with a few scenes and action:scene pairs in the validation are never seen in the training. 

\vspace{1mm}
\noindent\textbf{Qualitative Analysis.} To validate that \mtnet pays more attention to action cues and relevant objects than our baseline, we visualize our trained models using Grad-CAM ~\cite{Selvaraju_2019} to plot saliency maps for 6 videos in Fig. \ref{fig:saliency} and analyze where both networks attend to in each video.

We observe that \mtnet looks at regions more indicative of the action being performed. In column (a), while the baseline attends to the bowling alley environment and erroneously predicts \emph{bowling}, \mtnet attends to the subject's feet to recognize \emph{moonwalk}. In column (b), the baseline looks at the gymnasium environment and incorrectly predicts \emph{volleyball}, and \mtnet instead looks at the moving dancers to recognize \emph{high kick}. In column (c), the baseline attends to the beach environment and predicts \emph{beach soccer}, but \mtnet attends to the moving subject and recognizes \emph{tightrope walking}. In column (d), the baseline looks at the track and field scene and predicts \emph{javelin throw}, while \mtnet looks at the motion of the athletes and correctly predicts \emph{hurdling}. In column (e), \mtnet attends to the rake object and region of motion and correctly predicts the action of \emph{raking leaves}. Similarly, in column (f), \mtnet attends to the soccer object and region of motion and correctly predicts the action of \emph{passing soccer ball}.

\section{Conclusion}

We present Adversarial Multi-Task Neural Networks (\mtnet), a novel multi-task learning framework that generalizes existing works to include both auxiliary and adversarial tasks. This upends the common assumption that only beneficial auxiliary tasks should be included in multi-task learning. Our approach uses NCA as a data-driven first step toward the understanding of auxiliary and adversarial tasks for multi-task learning. Through this, we identify that scenes in HVU are merely correlated with actions and that it is of particular importance to address this scene bias. \mtnet subsequently improves action recognition on the challenging Scene-Invariant HVU splits. This approach can be generalized to any dataset with multiple class labels. 
An important future direction is to extend \mtnet beyond video understanding, where we might encounter tasks that cannot be classified as adversarial or auxiliary by NCA.

\noindent\textbf{Acknowledgements} This study was partially supported Panasonic and Stanford Institute for Human-Centered AI (HAI) Google Cloud Platform Credits.

%
%
\bibliographystyle{splncs04}
\bibliography{egbib}
\end{document}